\pdfoutput=1

\documentclass[11pt]{article}

\usepackage{ACL2023}

\usepackage{times}
\usepackage{latexsym}

\usepackage[T1]{fontenc}

\usepackage[utf8]{inputenc}

\usepackage{microtype}

\usepackage{inconsolata}

\usepackage{verbatim}
\usepackage{booktabs}
\usepackage{graphicx}
\usepackage{float}
\usepackage{amsmath}
\usepackage{multirow}
\usepackage{enumitem}

\title{Smart Word Suggestions for Writing Assistance}

\author{
      Chenshuo Wang\textsuperscript{1,2}\footnotemark[1],
      Shaoguang Mao\textsuperscript{3}\footnotemark[2],
      Tao Ge\textsuperscript{3},
      Wenshan Wu\textsuperscript{3},
      Xun Wang\textsuperscript{3}\\
      \textbf{
      Yan Xia\textsuperscript{3},
      Jonathan Tien\textsuperscript{3},  
      Dongyan Zhao\textsuperscript{1,2,4,5}} \\
    \textsuperscript{\rm 1}Wangxuan Institute of Computer Technology, Peking University\\
    \textsuperscript{\rm 2}Center for Data Science, Peking University
    \textsuperscript{\rm 3}Microsoft\\
     \textsuperscript{\rm 4}Institute for Artificial Intelligence, Peking University\\
     \textsuperscript{\rm 5}State Key Laboratory of Media Convergence Production Technology and Systems\\
\texttt{\{shaoguang.mao,wenshan.wu,tage,xunwang,yanxia,jtien\}@microsoft.com},\\\texttt{iven@ivenwang.com},  \texttt{zhaody@pku.edu.cn} \\}

\begin{document}
\maketitle
\footnotetext[1]{This work was performed during the first author’s internship at Microsoft Research Asia}
\footnotetext[2]{Corresponding Author}
\begin{abstract}

Enhancing word usage is a desired feature for writing assistance. To further advance research in this area, this paper introduces "Smart Word Suggestions" (SWS) task and benchmark.  Unlike other works, SWS emphasizes end-to-end evaluation and presents a more realistic writing assistance scenario. This task involves identifying words or phrases that require improvement and providing substitution suggestions. 
The benchmark includes human-labeled data for testing, a large distantly supervised dataset for training, and the framework for evaluation. The test data includes 1,000 sentences written by English learners, accompanied by over 16,000 substitution suggestions annotated by 10 native speakers. The training dataset comprises over 3.7 million sentences and 12.7 million suggestions generated through rules. 
Our experiments with seven baselines demonstrate that SWS is a challenging task. 
Based on experimental analysis, we suggest potential directions for future research on SWS. The dataset and related codes is available at \url{https://github.com/microsoft/SmartWordSuggestions}.
\end{abstract}

\section{Introduction}

\begin{figure}[ht!]
\centering
\includegraphics[scale=0.48]{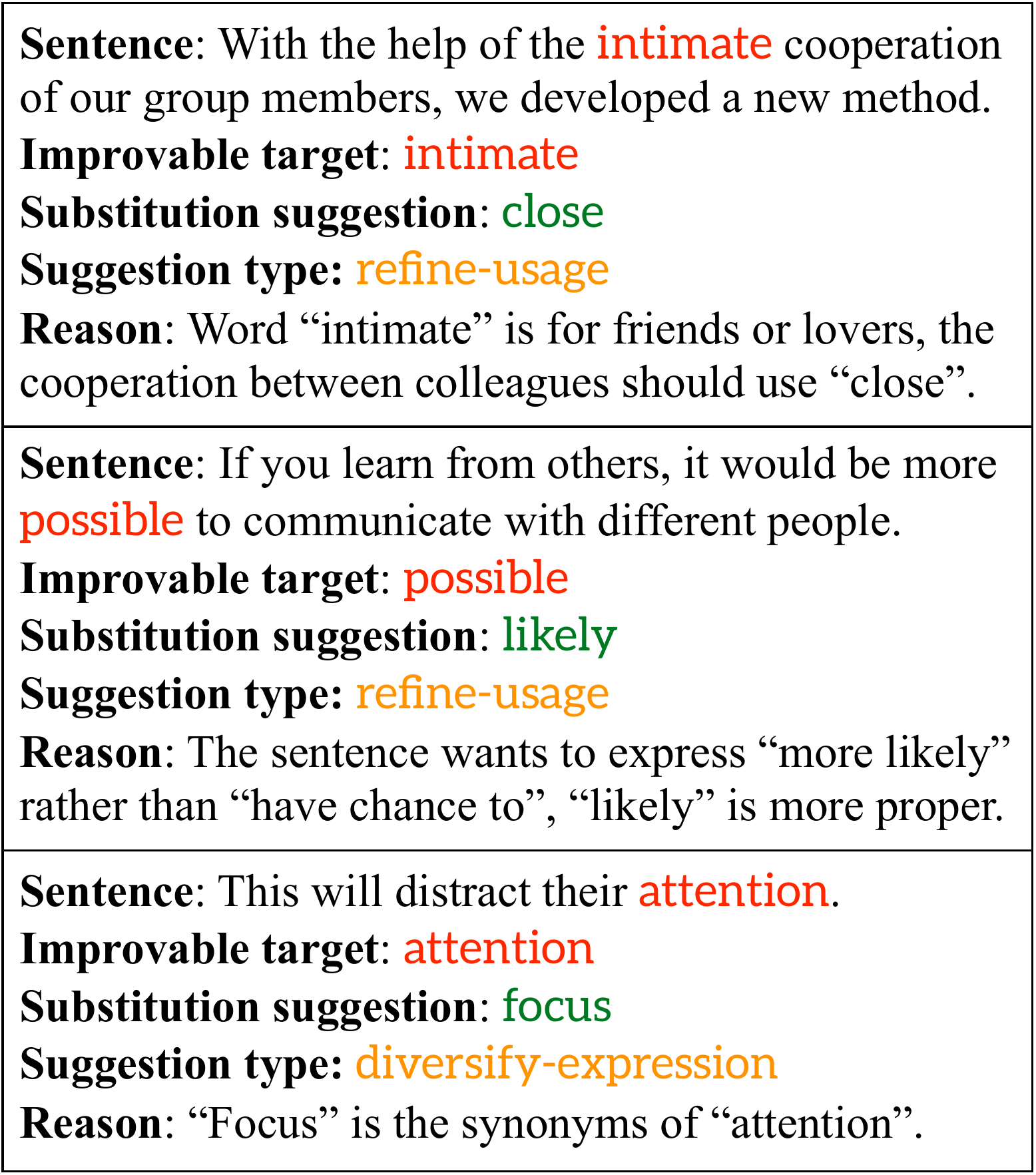}
\caption{
Examples for Smart Word Suggestions (SWS). 
All samples consist of sentences annotated with multiple improvable targets, 
each of which is further annotated with multiple substitution suggestions.
To save space, the sentences are simplified, 
and only one target and one suggestion are presented per case. 
The suggestions can be divided into two types: refine-usage and diversify-expression, 
which are described in section \ref{stage2}
}
\label{pic:swsexample}
\end{figure}

Writing assistance is a widely used application of natural language processing (NLP) that helps millions of people.
In addition to common features like grammatical error correction \cite{ng-etal-2014-conll, bryant-etal-2017-automatic}, 
paraphrasing \cite{fader-etal-2013-paraphrase, Lin2014MicrosoftCC} and 
automatic essay scoring \cite{ijcai2020p536}, providing word suggestions is a desired feature to enhance the overall quality of the writing.  As illustrated in figure \ref{pic:swsexample}, the word ``intimate'' in the first sentence should be replaced with ``close'', as ``intimate'' is not suitable for describing relationships between colleagues.

\begin{figure*}[ht]
\centering
\includegraphics[scale=0.45]{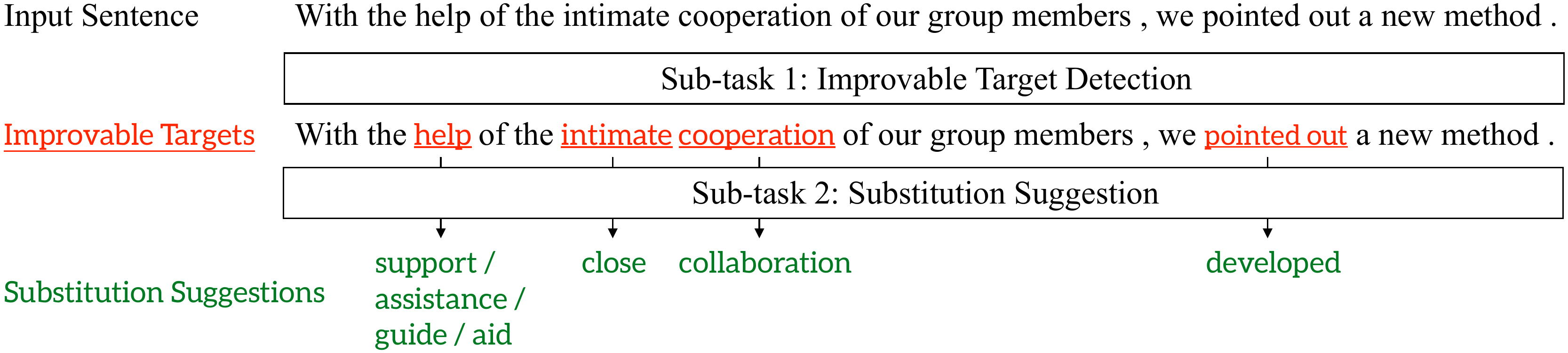}
\caption{
Task definition of Smart Word Suggestions (SWS).
SWS consists of two sub-tasks: improvable target detection and substitution suggestion. 
A sentence contains multiple improvable targets, 
and a target has multiple substitution suggestions. 
}
\label{pic:main}
\end{figure*}

In this paper, 
we introduce the task and benchmarks of \textbf{Smart Word Suggestion} (SWS). 
Figure \ref{pic:main} shows the  definition of SWS. 
The goal of SWS is to identify potential \textbf{improvable targets} in the form of words or phrases within a given context, 
and provide \textbf{substitution suggestions} for every improvable target. These suggestions may include correcting improper word usage, ensuring that language usage conforms to standard written conventions, enhancing expression, and so on. Specifically, we categorize these suggestions into two types: refine-usage and diversify-expression. 

Lexical Substitution (LS)  \cite{mccarthy-navigli-2007-semeval,kremer-etal-2014-substitutes,lee-etal-2021-swords} is the most relevant research benchmark in the field. 
LS systems aim to provide substitute words that maintain the original meaning of a given word within a sentence. 
However, in practical situations, it is important to recognize words that can be improved or replaced. Identifying these targets is crucial for practical use and a necessary step for making accurate substitution suggestions.
In order to reproduce the real-world scenarios, we design SWS as an end-to-end process that takes a sentence as input and provides substitution suggestions for all improvable targets as output. 

The SWS benchmark includes human-labeled data for testing, a large distantly supervised dataset for training, and a corresponding framework for evaluation. For testing, we collect 1,000 segments from English learners' essays, and ask ten annotators to identify improvable targets and provide substitution suggestions. The high level of agreement among the annotators confirms the quality of the annotation. For weakly supervised training, we compile a large amount of distantly supervised data by using a synonym thesaurus to randomly substitute words in corpus. We also provide settings for both end-to-end evaluation and sub-task evaluation.


To investigate the challenges, 
we implemented seven baselines, including knowledge-driven methods, state-of-the-art lexical substitution methods, and end-to-end approaches for SWS. 
The experimental results show that the performance of the existing lexical substitution methods decreases significantly when applied to SWS. 
Additionally, the end-to-end methods we designed struggle to identify and improve targeted words or phrases. Detailed analysis and discussions on the results suggest several areas for further research. 

To conclude, our contributions are as follows: 
\begin{itemize}[itemsep=2pt, topsep=0pt, parsep=0pt]
    \item Introducing the SWS task for writing assistance, and providing a benchmark with high-quality human-labeled testing data and large distantly supervised training data. 
    \item Developing the evaluation framework for SWS, and conducting extensive evaluations on the provided baselines. 
    \item Identifying several directions for further research on SWS through analysis. 
\end{itemize}


\section{Related Works}

We begin by comparing SWS with three related tasks,
highlighting the unique value of our work.   

\subsection{Lexical Substitution}

Lexical substitution  (LS) \cite{mccarthy-navigli-2007-semeval,kremer-etal-2014-substitutes,lee-etal-2021-swords} is the task of providing substitute words for a specific word in a sentence. 
There are some major distinctions between the SWS and LS.

(1) In LS, the target word is already provided, while in SWS, 
the system needs to detect the improvable targets first.

(2) LS focuses on finding synonyms that maintain the meaning of both the word and the sentence. On the other hand, SWS is designed for writing assistance scenarios, so the substitutions aim to improve the writing of the sentences.
LS focuses on word sense disambiguation in the context, which doesn't require any "improvement". Here is an example in the LS07 dataset:
\texttt{This is clearly a terrible and shameful blot on UN peacekeeping.}
One of the substitutions is "terrible" $\rightarrow$ "very bad". This substitution doesn't meet the SWS's requirement as the use of "very bad" is less accurate, and the substitution worsens writing. 

(3) LS uses lemmatized annotations for the target word and substitutions, while SWS extracts annotations directly from the sentence and requires that the substitutions fit grammatically within the sentence to evaluate the model's end-to-end performance.


\subsection{Grammatical Error Correction}

Grammatical error correction (GEC) \cite{ng-etal-2014-conll, bryant-etal-2017-automatic} also shares some similarities with SWS. 
\citet{ng-etal-2014-conll} pointed that more than 85\% of the corrections in GEC are word-level and that these corrections improve users' writing as well. 
However, the substitution suggestions provided by SWS do not include suggestions for correcting grammatical errors. 
Instead, SWS focuses on identifying and improving word or phrase usage. 
It is worth noting that the source sentences in the SWS test set are first processed by a GEC model \cite{ge-etal-2018-fluency} and then further checked by human annotators to ensure no grammatical errors in the inputs. 
In the writing assistant, SWS is the next step following GEC.

\subsection{Paraphrase Generation}

Paraphrase generation (PG) \cite{fader-etal-2013-paraphrase, Lin2014MicrosoftCC} aims to alter the form or structure of a given sentence while preserving its semantic meaning. 
PG has a variety of potential applications, such as data augmentation \cite{iyyer-etal-2018-adversarial}, query rewriting \cite{dong-etal-2017-learning}, and duplicate question detection \cite{shah-etal-2018-adversarial}. 
PG is different from SWS in two main ways: 
(1) SWS places a greater emphasis on improving writing by identifying and correcting inappropriate word usage or providing diverse expression options. 
(2) SWS focuses on substitution suggestions of words or phrases, 
and evaluations are based on word level. In contrast, PG directly measures performance at the sentence level.

\section{Data Collection}

This work is to construct a Smart Word Suggestion benchmark that accurately represents writing assistance scenarios. 
For evaluation, we collect sentences from English learners and use human annotations in accordance with \citet{mccarthy-navigli-2007-semeval} and \citet{kremer-etal-2014-substitutes}.
For training, 
we compile a large-scale, distantly supervised dataset from Wikipedia \cite{10.1007/978-3-319-11964-9_4, vrandevcic2014wikidata}. 

\subsection{Human-Annotated Data Collection} 

Human-annotated data is obtained through a three-stage process: 
(1) cleaning corpus data from English learners' essays, 
(2) labeling improvable targets and corresponding substitution suggestions, 
and (3) merging annotations and filtering out low-confidence annotations.


\paragraph{Stage 1: Corpus Cleaning.} 

We collect essays written by undergraduate English learners via an online writing assistance platform \footnote{\url{https://aimwriting.mtutor.engkoo.com/}}.
We divide them into individual sentences.
To avoid annotators making corrections beyond SWS, 
the sentences are refined with following actions:  
(1) removing sentences that have unclear meanings. 
(2) applying a correction model \cite{ge-etal-2018-fluency} to correct grammatical errors. 
(3) asking human reviewers to double-check for any remaining grammatical errors.
Additionally, 
we filter out short sentences as they may not provide enough context or contain sufficient words to improve. 
We thoroughly reviewed all sentences to ensure that they do not contain any information that could identify individuals or any offensive content.

\paragraph{Stage 2: Human Annotation.} \label{stage2}

Ten native English-speaking undergraduate students majoring in linguistics were recruited as annotators to independently annotate each sentence. 
To ensure annotation quality, 
all annotators were required to pass test tasks before participating in the annotation.

The annotators carried out the annotations in three steps: 
(1) identifying words or phrases in the sentence that could be improved, 
(2) offering one or more suggestions for each identified target, 
and (3) assigning a type of improvement after the substitution. 

Specifically, 
we define the substitution suggestions as two types. 
(1) \textbf{Refine-usage} refers to instances where the use of a specific word or phrase is inappropriate in the current context, such as when it has a vague meaning, is a non-native expression, or is an incorrect usage of English.
For instance, in the second sentence shown in figure \ref{pic:swsexample}, the word "possible" is intended to convey the meaning of "having the possibility", and is not appropriate in the context of the sentence. The annotators replaced "possible" with "likely." These suggestions are designed to help English learners understand the differences in word usage in specific contexts and to enable them to write in a way that is more consistent with native speakers.
(2) \textbf{Diversify-expression} refers to instances where this word or phrase could be substituted with other words or phrases.
These suggestions aim to help users use a more diverse range of expressions. 
The last case in figure \ref{pic:swsexample} is a corresponding example.

The annotators were required to provide at least three suggestions for each sentence. 
For the entire dataset of 1000 sentences, 
each annotator was required to provide at least 1500 refine-usage type suggestions. 
The detailed annotation instruction is in appendix \ref{annotateguide}.

\paragraph{Stage 3: Merging and Filtering.} \label{annotates3}

Previous lexical substitution tasks \cite{mccarthy-navigli-2007-semeval, kremer-etal-2014-substitutes} 
merged all the annotators' results into a key-value dictionary, 
where the value indicates the number of annotators who provided this substitution suggestion. We merged the labeling results of 10 annotators in a similar way.  
Take the merging of two annotators' annotations as an example.
One is 
\texttt{\{happy: glad/merry, possible: likely\}}, 
and the other is \texttt{\{help: aid, possible: likely/probable\}}. 
The result after merging would be:

\texttt{    \{happy: \{glad: 1, merry: 1\},}

\texttt{    possible: \{likely: 2, probable: 1\},}

\texttt{    help: \{aid: 1\}\}}


where \texttt{happy, possible, help} are improvable targets, 
and the sub-level dictionaries are the substitution suggestions after merging. 
We also collect the type of refine-usage or diversify-expression for each improvable target by taking the majority of the type labeling.

In order to reduce subjective bias among annotators, 
we discarded all improvable targets that were only annotated by one annotator. 
Finally, the dataset was split into a validation set of 200 sentences and a test set of 800 sentences. 

\subsection{Distantly Supervised Data Collection} \label{distantsupervise}

We collect a large amount of distantly supervised data for weakly supervised training by using a synonym thesaurus to randomly substitute words in a corpus.
The source corpus contains 3.7 million sentences from Wikipedia\footnote{\url{https://dumps.wikimedia.org/enwiki/20220720/}} . 
The synonym thesaurus we use is the intersection of PPDB \cite{pavlick-etal-2015-ppdb} and Merriam-Webster thesaurus\footnote{\url{https://www.merriam-webster.com/thesaurus}}. 
The sentences are processed in 3 steps: 
(1) Selecting all the words or phrases in the synonym thesaurus, 
and treating them as improvable targets. 
(2) Using a tagger to find the part of speech of the improvable targets. 
(3) Randomly substituting the improvable targets with one synonyms of the same part of speech. 

Note that the random substitution with the synonym dictionary may result in a more inappropriate word or phrase usage than the original text. 
Therefore, we treat the generated substitutions as the improvable targets, and the original targets as substitution suggestions.  

In contrast to the human-annotated dataset, the distantly supervised dataset only includes one suggestion for each improvable target and does not have the annotation of suggestion type. 
The code for generating distantly supervised datasets will be released for further studies. 

\subsection{Data Statistics} \label{datastats}

\begin{table}[h]
\centering
\small
\setlength \tabcolsep{1.3pt}
\begin{tabular}{@{}lrrrr@{}}
\toprule
Benchmark       & \# Sentence  & \# Target     & \# Suggestion  & \# Label     \\ \midrule
SemEval         & 2010      & 2010       & 8025        & 12,300     \\
\textsc{CoInCo} & 2474      & 15,629     & 112,742     & 167,446    \\
\textsc{Swords} & 1250      & 1250       & 71,813      & 395,175    \\ \midrule
SWS             & 1000      & 7027       & 16,031      & 30,293     \\
SWS$_{DS}$      & 3,746,142 & 12,786,685 & 12,786,685  & 12,786,685 \\ \bottomrule
\end{tabular}
\caption{
Statistics of SWS and LS datasets. 
SWS$_{DS}$ stands for the distantly supervised dataset. 
}
\label{tab:stat}
\end{table}

Table \ref{tab:stat} shows the comparison between SWS and lexical substitution benchmarks. Our SWS dataset consists of 7027 instances of improvable targets and 16031 suggestions in 1000 sentences. The average length of the sentences in this dataset is 27.8 words.
The improvable targets in this dataset includes 2601 nouns, 2186 verbs, 1263 adjectives, 367 adverbs, 267 phrases, and 343 other parts of speech. 3.8\% of the targets and 3.3\% of the suggestions are multi-word phrases. 
63.0\% of the targets are the type of refine-usage. Table \ref{tab:statpos} shows the proportion of refine-usage or diversify-expression targets with different part-of-speech. 

\begin{table}[h]
\small
\setlength \tabcolsep{4.1pt}
\begin{tabular}{@{}llllllll@{}}
\toprule
POS     & noun & verb & adj. & adv. & phrase & others & total \\ \midrule
number  & 2601 & 2186 & 1263 & 367  & 267    & 343    & 7027  \\
RU (\%) & 57.8 & 63.7 & 66.7 & 64.9 & 70.8   & 76.7   & -     \\
DE (\%) & 42.2 & 36.3 & 33.3 & 35.1 & 29.2   & 23.3   & -     \\ \bottomrule
\end{tabular}
\caption{
Statistics of targets with different part-of-speech. 
RU refers to the proportion of refine-usage targets, 
and DE refers to the proportion of diversify-expression. 
}
\label{tab:statpos}
\end{table}

The distantly supervised dataset SWS$_{DS}$ contains over 12.7 million suggestions in 3.7 million sentences. 2.67\% are multi-word phrases, and 0.3\% of the suggestions are multi-word.







\subsection{Inner Annotator Agreements}

Previous studies on lexical substitution\cite{mccarthy-navigli-2007-semeval, kremer-etal-2014-substitutes} evaluated the quality of the dataset with inter-annotator agreement (IAA). 
We adopt this approach and calculate pairwise inter-annotator agreement (PA) to assess the quality of the dataset.

$\mathbf{{PA}^{det}}$ measures the consistency of identifying improvable targets: 
$$\mathbf{{PA}^{det}} = \frac{1}{|P|}\sum_{ (i, j) \in P} \mathbf{{PA}^{det}}_{ij}$$
$$\mathbf{{PA}^{det}}_{ij} = \sum_{k=1}^N \frac{1}{ N }\frac{|s^i_k 
\cap s^j_k|}{|s^i_k \cup s^j_k|} $$
\noindent where $P$ is the set of annotator pairs. 
We have ten annotators, so $|P|=C_{10}^2=45$. 
$N$ is the number of all the sentences, 
and $s^i_k, s^j_k$ are the improvable target sets of sentence $k$ identified by annotator $i$ and $j$, respectively.

$\mathbf{{PA}^{sug}}$ measures the consistency of substitution suggestions of a same improvable target: 
$$\mathbf{{PA}^{sug}} = \frac{1}{|P|} \sum_{( i, j) \in P} \mathbf{PA}^{sug}_{ij}$$
$$\mathbf{{PA}^{sug}}_{ij} = \sum_{l=1}^{M_{ij}} \frac{1}{M_{ij}} \frac{|t^i_l
\cap t^j_l|}{|t^i_l \cup t^j_l|} $$
\noindent where $M_{ij}$ is the size of the intersection of the improvable target sets identified by annotator $i$ and $j$. 
$t^i_l, t^j_l$ are the suggestions for target $l$ given by annotator $i$ and $j$, respectively.


In the SWS benchmark, 
the $\mathbf{PA}^{det}$ and the $\mathbf{PA}^{sug}$ are 23.2\% and 35.4\%, respectively. 
Our $\mathbf{PA}^{sug}$ is significantly higher compared to previous LS datasets, 27.7\% of SemEval \cite{mccarthy-navigli-2007-semeval} and 19.3\% of 
\textsc{CoInCo} \cite{kremer-etal-2014-substitutes}, thereby confirming the annotation quality.

\subsection{Data Quality of the Distantly Supervised Dataset}

According to our statistics, 71.8\% of the substitutions in the test set appear in the training set, and each substitution in the test set appears in the training set 10.4 times on average. Those data show the substitutions in the training set covers most of the substitutions in the test set, which verify the synthetic method is close to real-world scenarios.

\section{Evaluation}

In this section, 
we introduce the evaluation settings and metrics for SWS, 
including both the end-to-end evaluation and the sub-task evaluation. 

For the end-to-end evaluation and the improvable target detection sub-task, 
we introduce precision, recall, and F$_{0.5}$ as metrics. 
For the substitution suggestion sub-task, 
we utilize accuracy to evaluate the quality of the predicted substitutions. Examples of calculating the metrics can be found in appendix \ref{evalexample}. 

\subsection{End-to-end Evaluation}

The end-to-end evaluation is computed based on each substitution suggestion. 
A true prediction is counted if and only if both the detected improvable target is in the annotated improvable target set and the suggested substitution is in the annotated substitutions of the target:  
$$\mathbf{TP^{e2e}} = \sum_{k=1}^{N}\sum_{l=1}^{M_k} 1 \text{\ if\ } s_{kl} \in S_k \text{\ else\ } 0$$

\noindent where $N$ is the number of all the sentences, 
$M_k$ is the number of targets in the sentence $k$, 
$S_k$ is the set of annotated suggestions of sentence $k$, 
and $s_{kl}$ is the $l$-th predicted suggestion of sentence $k$. 
The precision ($\mathbf{P^{e2e}}$) and recall ($\mathbf{R^{e2e}}$) for end-to-end evaluation are calculated as follows: 
$$\mathbf{P^{e2e}} = \frac{\mathbf{TP^{e2e}} }{ N_P }, \ \mathbf{R^{e2e}} = \frac{\mathbf{TP^{e2e}} }{ N_G }$$

\noindent where $N_P$ and $N_G$ are the number of predicted suggestions and annotated suggestions, 
respectively. 
In the writing assistance scenario, 
precision is more important than recall, 
so we calculate $\mathbf{F_{0.5}^{e2e}}$ as the overall metric. 
$$\mathbf{F_{0.5}^{e2e}} = \frac{ 1.25 \cdot \mathbf{P^{e2e}} \cdot \mathbf{R^{e2e}} }{ 0.25 \cdot \mathbf{P^{e2e}} + \mathbf{R^{e2e}} }$$

\subsection{Sub-Task Evaluation}

\paragraph{Improvable Target Detection.}

In this task, model needs to find all the annotated improvable targets in the sentence. The precision ($\mathbf{P^{det}}$) and recall ($\mathbf{R^{det}}$) for detection are calculated as follows:
$$\mathbf{P^{det}} = \frac{\sum_{k=1}^N |s_k \cap s'_k | }{\sum_{k=1}^N | s'_k |} , \  \mathbf{R^{det}} = \frac{\sum_{k=1}^N |s_k \cap s'_k | }{\sum_{k=1}^N | s_k |}$$

\noindent where $s_k$ and $s'_k$ are the annotated improvable target set and predicted improvable target set for sentence $k$, respectively. 
Same with end-to-end evaluation, 
we compute $\mathbf{F_{0.5}^{det}}$ to assess the performance for detection of improvable targets.
$$\mathbf{F_{0.5}^{det}} = \frac{ 1.25 \cdot \mathbf{P^{det}} \cdot  \mathbf{R^{det}} }{ 0.25 \cdot \mathbf{P^{det}} + \mathbf{R^{det}} }$$

\paragraph{Substitution Suggestion.}

In this task, model needs to give suggestions for each improvable target. 
We calculate accuracy of the suggestions on those correctly detected targets:
$$\mathbf{Acc^{sug}} = \frac{1}{N} \sum_{k=1}^N \bigg(\frac{1}{M_k} \sum_{l=1}^{M_k} 1 \text{\ if\ } t'_l \in T_l \text{\ else\ } 0 \bigg) $$

where $T_l$ is the annotated recommendation set of target $l$, 
$t'_l$ is the predicted recommendation for target $l$, 
and $M_k$ is the total number of correctly detected targets in sentence $k$.

\section{Experiments}

\begin{table*}[h]
\centering
\begin{tabular}{@{}ll|rclc|lll@{}}
\toprule
 &  & \multicolumn{4}{c|}{Sub-task Evaluation} & \multicolumn{3}{l}{End-to-end Evaluation} \\ \cmidrule(l){3-9} 
 & Model & $\mathbf{P^{det}}$ & $\mathbf{R^{det}}$ & \multicolumn{1}{r|}{$\mathbf{F^{det}_{0.5}}$} & $\mathbf{Acc^{sug}}$ & $\mathbf{P^{e2e}}$ & $\mathbf{R^{e2e}}$ & $\mathbf{F^{e2e}_{0.5}}$ \\ \midrule
External Knowledge & Rule-based & 0.585 & 0.344 & \multicolumn{1}{r|}{0.513} & 0.314 & 0.183 & 0.108 & 0.161 \\
Methods & ChatGPT & 0.451 & 0.418 & \multicolumn{1}{r|}{0.444} & 0.427 & 0.193 & 0.179 & 0.190 \\ \midrule
Lexical Substitution & BERT$_{s_p, s_v}$ & 0.511 & 0.050 & \multicolumn{1}{r|}{0.180} & 0.441 & 0.225 & 0.022 & 0.079 \\
Methods & LexSubCon & 0.438 & 0.667 & \multicolumn{1}{r|}{0.470} & 0.281 & 0.123 & 0.188 & 0.132 \\ \midrule
\multirow{3}{*}{\begin{tabular}[c]{@{}l@{}}End-to-End\\ Methods\end{tabular}}  & CMLM & 0.512 & 0.222 & \multicolumn{1}{r|}{0.406} & 0.236 & 0.121 & 0.052 & 0.096 \\
 & BART & 0.555 & 0.243 & \multicolumn{1}{r|}{0.441} & 0.446 & 0.248 & 0.108 & 0.197 \\
 & BERT & 0.585 & 0.249 & \multicolumn{1}{r|}{0.460} & 0.436 & 0.255 & 0.108 & 0.201 \\ \midrule \midrule
& Human* & 0.709 & 0.313 & \multicolumn{1}{r|}{0.566} & 0.631 & 0.449 & 0.199 & 0.359 \\
 \bottomrule
\end{tabular}
\caption{
Evaluation results on SWS. 
*: As a reference, we offer human performance by taking the average of ten rounds of evaluations. In each round, each annotator is compared to the combined annotations of other annotators.}
\label{tab:mainres}
\end{table*}

\subsection{Baselines}
We test 7 methods on SWS. 
The methods could be divided into three groups: 
(1) Adopting external knowledge to give suggestions. 
(2) State-of-the-art lexical substitution methods. 
(3) End-to-end SWS baselines. We also list the human performance for reference.

\paragraph{External Knowledge Methods.}

Here are two methods that use external knowledge to give suggestions. 
(1) Rule-based synonyms replacement as how we construct the distantly supervised data. 
We adopt a greedy replacement strategy, where all entries are replaced.
(2) ChatGPT\footnote{\url{https://openai.com/blog/chatgpt/}}, 
a large language model trained on massive data and further fine-tuned with human feedback.
We ask ChatGPT to directly generate the suggestions in every giving sentence. 
The prompt and details for utilizing ChatGPT can be found in appendix \ref{chatgptprompt}.

\paragraph{Lexical Substitution Methods.}

Two state-of-the-art lexical substitution methods are tested on SWS, 
i.e. BERT$_{s_p, s_v}$ \cite{zhou-etal-2019-bert} and LexSubCon \cite{michalopoulos-etal-2022-lexsubcon}. 
We use the open-sourced code of LexSubCon and re-implement BERT$_{s_p, s_v}$. 
We let the model give a substitution for each word, and if the substitution is different with the original word, the word is regarded as a detected improvable target. 

\paragraph{End-to-end Baselines.}

In the end-to-end framework, we treat SWS as three training paradigms, 
and provide one baseline for each. 
(1) Masked language modeling (MLM): 
We use BERT-base-uncased \cite{devlin-etal-2019-bert} with an MLM head as the baseline. 
(2) Sequence-to-sequence generation: 
We use BART-base \cite{lewis-etal-2020-bart} as the baseline. 
(3) Token-level rewriting: 
We use CMLM \cite{ghazvininejad2019MaskPredict} as the baseline. 
The distantly supervised dataset is utilized to train the end-to-end baselines. 
For the improvable targets, the model is expected to learn the suggestions. Otherwise, the model is expected to keep the original words.

\subsection{Main Results} \label{mainres}

Table \ref{tab:mainres} shows the experimental results of the baselines, 
from which we have the following observations: 

(1) The rule-based approach is similar to the process of creating distantly supervised data. Both the rule-based method and end-to-end baselines, which are trained using distantly supervised data, have high $\mathbf{P^{det}}$ and low $\mathbf{R^{det}}$ values. This suggests that the synonym dictionary used in this work has high quality but low coverage.

(2) Compared with the rule-based method, 
the end-to-end models trained on distantly supervised dataset show a decline in performance for the improvable target detection, but an increase in performance for substitution suggestion.
The improvable targets of the distantly supervised data do not accurately reflect the words or phrases that need improvement, resulting in difficulty in effectively training the models in detecting. 
However, the substitution suggestions in the distantly supervised data are derived from original words in Wikipedia, enabling the models to learn a relatively appropriate word usage in context.

(3) The results of the CMLM model show a decrease in performance compared to the pre-trained models, namely BERT and BART, particularly in terms of substitution suggestions. The pre-training of semantic knowledge may contribute to the superior performance of the pre-trained models for this task.

(4) There is a notable decrease in SWS for LS methods. Moreover, different LS methods have significant differences in detecting improvable targets.
Only 2.1\% of the words in the input sentence are identified as improvable targets by BERT$_{s_p,s_v}$, while LexSubCon detects 32.4\%. The current LS methods are not compatible with the SWS task.

(5) The results from ChatGPT are comparable with the end-to-end baselines trained on 3.7 million sentences, but it is still has room for improvement.

(6) Human performance is significantly better than baselines. We believe there is a lot of room for the baselines to improve.

\section{Analysis}

We analyze the experimental results with two questions: 
(1) Does the model have the capability to accurately identify words that require improvement, or does it simply make random guesses? 
(2) Does the model have the ability to provide multiple useful suggestions for each target word?

\subsection{Detection Analysis}

\paragraph{Voting Index and Weighted Accuracy. }

\begin{figure}[h]
\centering
\includegraphics[scale=0.8]{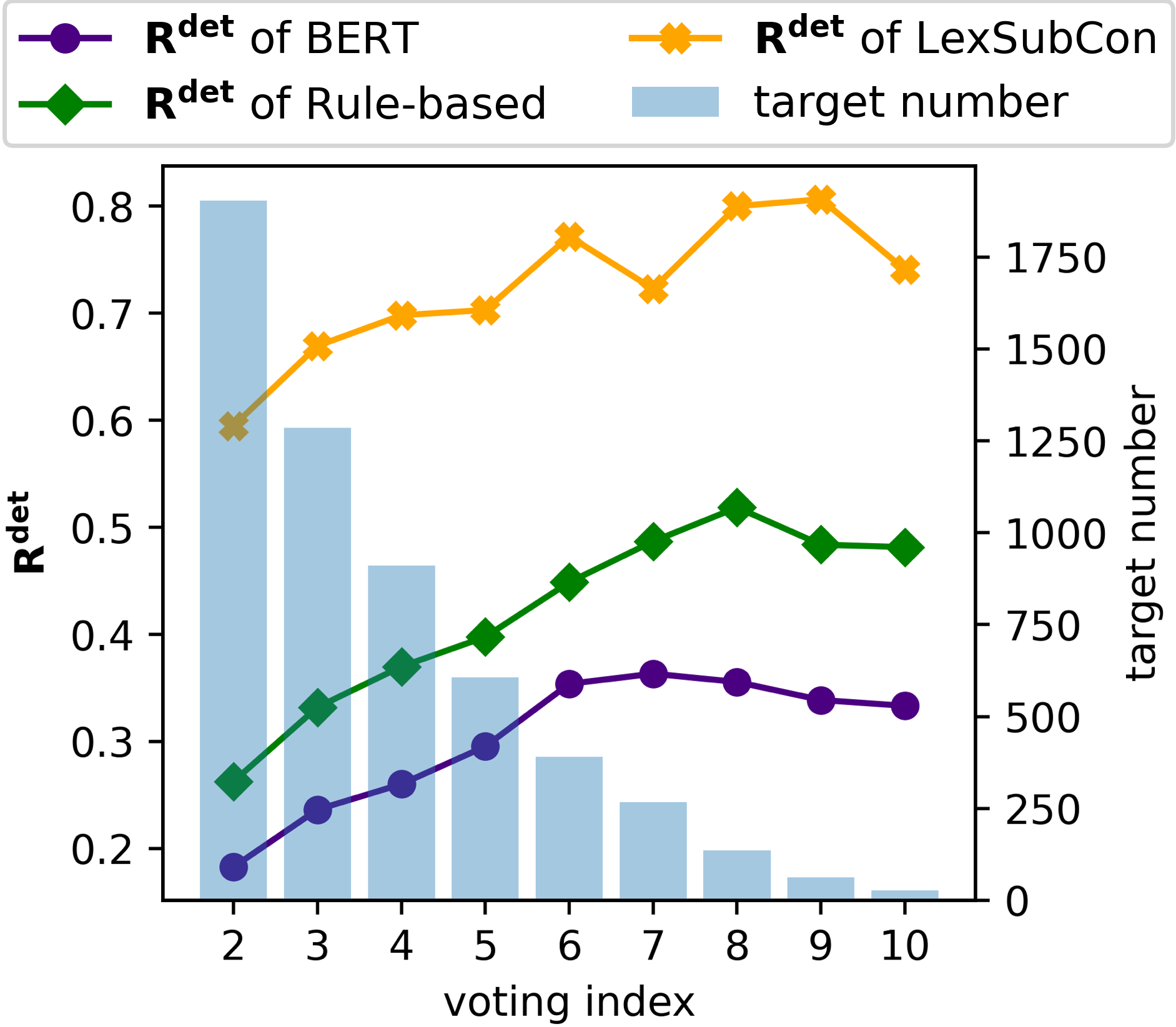}
\caption{
The number of targets and $\mathbf{R^{det}}$ on different voting index. 
}
\label{pic:recall_improvable_level}
\end{figure}

After merging the annotations, we determine the voting index for each improvable target, 
i.e. the number of annotators who identified the word or phrase. 
The voting index reflects the necessary level of replacement for the word. 
Figure \ref{pic:recall_improvable_level} shows $\mathbf{R^{det}}$ for the improvable targets with different voting indexes. 
As depicted in Figure \ref{pic:recall_improvable_level}, 
improvable targets identified by a greater number of annotators are more easily detected by the models.

Then, we design weighted accuracy (WA) to evaluate the detection performance, using the voting index as weighting factors.

\vspace{-15pt}
$$\mathbf{WA^{det}} = \frac{\sum_{k=1}^N \sum_{l=1}^{M_k} w_{kl} \text{ if } s_{kl} \in s'_k \text{ else }0}{\sum_{k=1}^N \sum_{l=1}^{M_k} w_{kl}}$$

\noindent where $s'_{k}$ is the predicted improvable target set of sentence $k$, 
$s_{kl}$ is the $l$-th annotated target in sentence $k$, 
$w_{kl}$ is the voting index of $s_{kl}$, 
$N$ is the number of total sentences, 
and $M_k$ is the size of annotated improvable target set of sentence $k$. 

\begin{table}[h]
\centering
\begin{tabular}{@{}lrrr@{}}
\toprule
Model                 & ImpR  & $\mathbf{R^{det}}$ & $\mathbf{WA^{det}}$ \\ \midrule
Rule-based            & 0.125 & 0.344 & 0.382 \\ 
ChatGPT               & 0.224 & 0.418 & 0.449 \\ \midrule
BERT$_{s_p, s_v}$     & 0.021 & 0.050 & 0.061 \\ 
LexSubCon             & 0.324 & 0.667 & 0.694 \\ \midrule
CMLM                  & 0.094 & 0.222 & 0.239 \\
BART                  & 0.102 & 0.243 & 0.272 \\
BERT                  & 0.093 & 0.249 & 0.278 \\ \midrule
Human                 & 0.212 & -     & -     \\ \bottomrule
\end{tabular}
\caption{
Improvable ratio (ImpR), Detection Recall ($\mathbf{R^{det}}$) and Weighted Accuracy (WA) for improvable targets detection on SWS benchmark sets.
}
\label{tab:wares}
\end{table}

Table \ref{tab:wares} shows $\mathbf{R^{det}}$ and $\mathbf{WA^{det}}$ of baseline methods. 
Consistent with the trend of $\mathbf{R^{det}}$ for different voting indexes, 
the $\mathbf{WA^{det}}$ is relatively higher than $\mathbf{R^{det}}$. 
These results demonstrate that the baseline methods can detect the high-confidence improvable targets better. 

\begin{table*}[]
\centering
\begin{tabular}{@{}lllrllrrrr@{}}
\toprule
Dataset  & $\mathbf{P^{det}}$ & $\mathbf{R^{det}}$ & $\mathbf{F^{det}_{0.5}}$ & $\mathbf{ImpR}$ & $\mathbf{WA^{det}}$ & $\mathbf{Acc^{sug}}$ & $\mathbf{P^{e2e}}$ & $\mathbf{R^{e2e}}$ & $\mathbf{F^{e2e}_{0.5}}$ \\ \midrule
Wiki-13.2\%              & 0.585 & 0.249 & 0.460 & 0.093 & 0.278 & 0.436 & 0.255 & 0.108 & 0.201  \\
Wiki-25.4\%              & 0.568 & 0.354 & 0.506 & 0.136 & 0.402 & 0.243 & 0.138 & 0.086 & 0.123  \\ \bottomrule
\end{tabular}
\caption{
Comparison of BERT trained on two distantly supervised datasets. 
The suffix stands for the constructed improvable target ratio of the dataset. 
The model trained on the dataset with more improvable targets yields a higher ImpR and a higher $\mathbf{R^{det}}$, but a worse performance in substitution suggestions.}
\label{tab:imprCompare}
\end{table*}

\paragraph{Improvable Ratio.} 
The improvable ratio (ImpR) is defined as the proportion of the number of detected improvable words to the total number of words in sentences. 
As shown in Table \ref{tab:wares}, 
$\mathbf{R^{det}}$, $\mathbf{WA^{det}}$ are positively correlated with ImpR. 

To investigate how to control the model to achieve a desired ImpR, we build another distantly supervised dataset for training. 
Different from dataset construction described in section \ref{distantsupervise}, 
we use the union of PPDB \cite{pavlick-etal-2015-ppdb} and Merriam-Webster thesaurus as a large synonym thesaurus. As the thesaurus size increases, the artificial improvable targets in constructed data are increased to 25.4\% from 13.2\%.

The results of BERT trained on two datasets are presented in Table \ref{tab:imprCompare}. Upon comparison of the two experiments, it is observed that the number of constructed improvable targets in the training set is nearly doubled, while the ImpR of the trained models only increases to 13.6\% from 9.3\%. It is challenging to control the ImpR.
Thus, one direction under research is to control the model to attain a desired ImpR while maintaining a good performance.




\subsection{Multiple Suggestions Analysis}

It may be beneficial for users to have multiple suggestions for each improvable target. Therefore, we design a multiple-suggestion setting that allows the system to provide multiple substitution suggestions for each detected improvable target.

As the output suggestions are ranked in order, we propose using Normalized Discounted Cumulative Gain ($\mathbf{NDCG}$), a metric commonly used in search engines, to measure the similarity between a ranked list and a list with weights.
$$\mathbf{NDCG}_m = \frac{1}{M} \sum_{k=1}^{M} \frac{\mathbf{DCG}_m(\text{T}_k')}{\mathbf{DCG}_m(\text{T}_k)}$$
$$\mathbf{DCG}_m(\text{T}_k) = \sum_{i=1}^m \frac{\sum_{i'\leq i} w_i}{\log(1 +i)}$$
$$\mathbf{DCG}_m(\text{T}'_k) = \sum_{j=1}^m \frac{\sum_{j'\leq j} w_j'}{\log(1 + j)}$$
$$w_j' =  w_i \ \text{ if } t'_{kj} \in T_{k}\ \text{ else } 0$$
In this formula, $M$ is the total number of true predicted improvable targets, and $m$ is a parameter that specifies the number of suggestions for an improvable target. 
In the numerator, we accumulate the weights for the predicted suggestions from the first to the last.
If recommendation $i'$ is not in human annotation, 
the weight is set to zero. Otherwise, the weight is set to its voting index. 
The denominator is a list sorted according to the voting index, 
which represents the optimal condition for giving $m$ predictions. 
We provide an example of calculating NDCG in appendix \ref{ndcgexample}.


\begin{figure}[]
\centering
\includegraphics[scale=0.8]{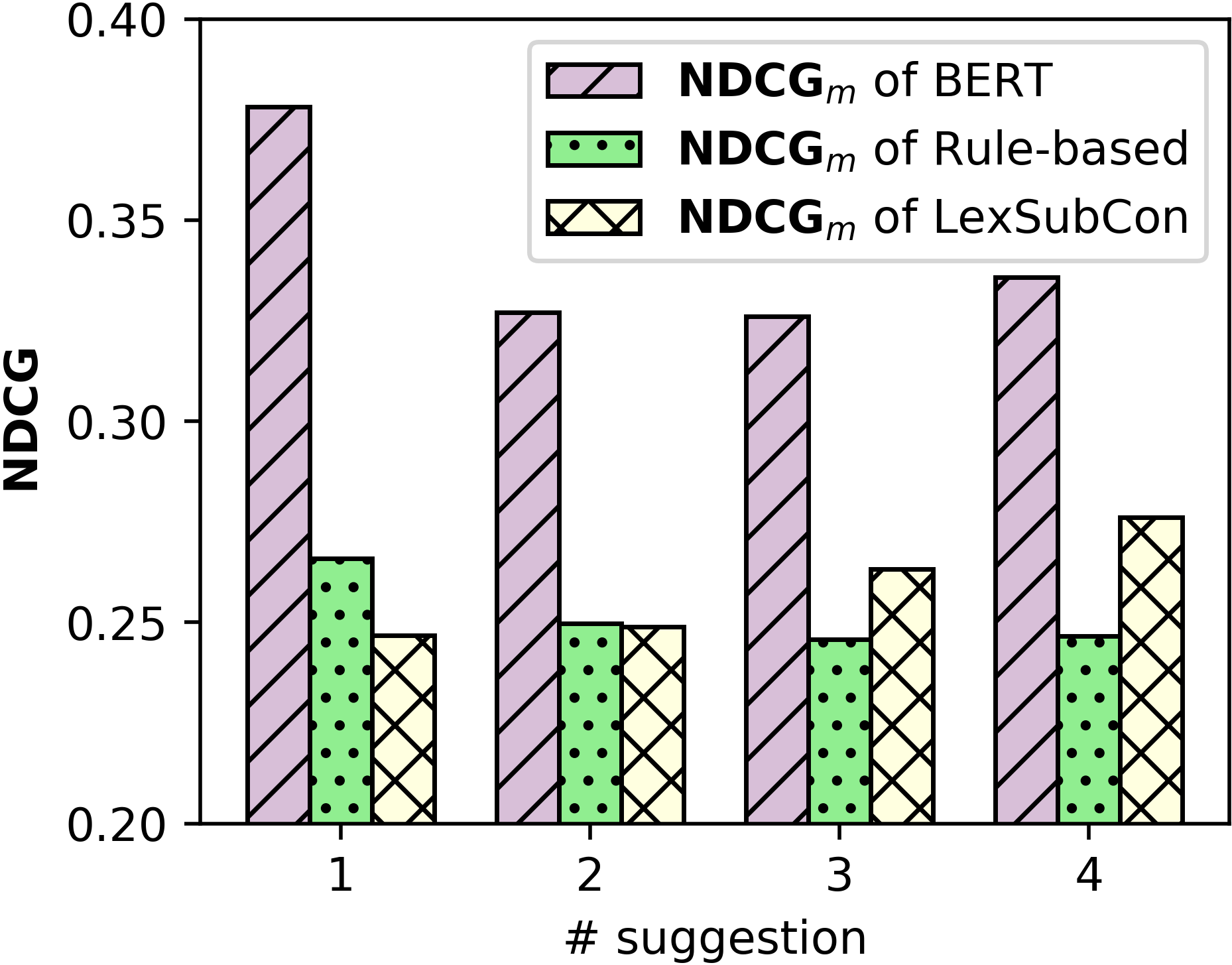}
\caption{
$\mathbf{NDCG}_m$ on different $m$ of BERT, rule-based method, and LexSubCon. 
}
\label{pic:ndcg}
\end{figure}

\begin{figure*}[]
\centering
\includegraphics[scale=0.5]{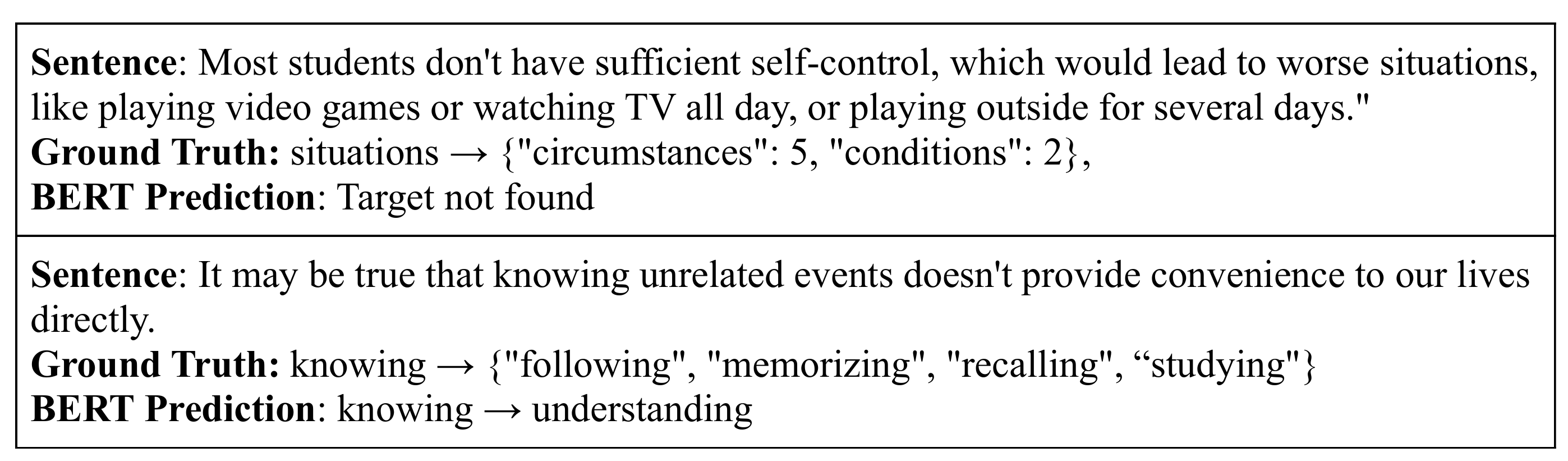}
\caption{
Case study of the BERT's predictions. 
}
\label{pic:case}
\end{figure*}

The average number of substitution suggestions for each improvable target in SWS benchmark is 3.3. When $m$ exceeds the substitution number for a given target, 
$\mathbf{DCG}_m(\text{T}_k)$ remains constant. Thus, $\mathbf{NDCG}_m$ is only calculated for $m=1,2,3,4$. Figure \ref{pic:ndcg} lists $\mathbf{NDCG}_m$ for different baselines.

BERT may perform better than other methods, but as the number of suggestions $m$ increases, the $\mathbf{NDCG}_m$ of BERT drops significantly. This suggests that BERT struggles when providing multiple suggestions. This could be due to the lack of multiple substitution suggestions in the distantly supervised dataset. Future research could focus on improving the model's ability to provide multiple substitution suggestions.





\subsection{Case Study}

Figure \ref{pic:case} gives two cases of BERT's predictions. 
In the first case, 
BERT didn't detect this improvable target. However, in our distantly-supervised training data, there are dozens of cases substituting "situations" to "circumstances". We think controlling the initiative of detecting is a direction worthy of research.

In the second case, 
BERT give the suggestion of "understanding", which is the closest word to "knowing" if ignores the context. However, it's not the right meaning in the context of "knowing events". 
We think it's hard to train a model aware of word usage in different contexts with the current distantly-supervised training data. Because we think the one-substitute-one data doesn't provide enough information for model training on word usage. We regard this as a future research direction.

\section{Conclusion}
This paper introduces the first benchmark for Smart Word Suggestions (SWS), 
which involves detecting improvable targets in context and suggesting substitutions. Different from the previous benchmarks, SWS presents a more realistic representation of a writing assistance scenario. 
Our experiments and analysis highlight various challenges for future research and suggest opportunities for improvement in future work.
We encourage further research on building more realistic training data, designing better data augmentation strategies, and developing unsupervised or self-supervised methods for SWS. 

\section{Limitations}

The SWS benchmark have two limitations: 
(1) The sentences in the SWS testing set come from students' essays, which limits the system's ability to test its performance in other specific domains such as laws or medicine. 
(2) the SWS corpus is at the sentence level, but some writing suggestions can only be made after reading the entire article, which are not included in our SWS dataset.

\bibliography{custom}
\bibliographystyle{acl_natbib}

\appendix

\section{Annotation Instructions} \label{annotateguide}

We need to find at least 3 "words/phrases to change" in a sentence, and give "substitutes" for each. Every substitute should be classified as improve-usage or diversify-expression. 

\subsection{What is the word/phrase that needs to change?}

Our aim is to find a word/phrase that needs to be better in writing scenarios. Suppose you are the teacher, and now you are helping the language learners to improve their English writing. We define a "word to change" as the substitution has influences as follows:

\begin{itemize}[itemsep=2pt, topsep=0pt, parsep=0pt]
\item To express the original semantic meaning more appropriately. 
\item To make the usage of the word much closer to the native speaker. 
\item To change spoken language into written language. 
\item To diversify the word usage for better expression. 
\end{itemize}

The substitution should NOT cause the influence as follows:

\begin{itemize}[itemsep=2pt, topsep=0pt, parsep=0pt]
\item Rewrite the sentence, instead of words or phrases, into a better expression (e.g. "it is advisable" → "advisably,").
\item Correct the mistakes in the sentence (e.g. "a lot" → "a lot of" in the sentence "There are a lot of valuable tips").
\item Substitute the word with a synonym, but not help the English learners with better writing.
\end{itemize}

After the definition, we also give some rules that you could refer to:

\begin{itemize}[itemsep=2pt, topsep=0pt, parsep=0pt]
\item the word/phrase that needs to change is usually less than 3 words.
\item the word/phrase that needs to change is usually an adj./adv./noun/verb.
\item the word/phrase that needs to change is usually not a named entity.
\end{itemize}

\subsection{How to give the substitutions?}

The substitution should:

\begin{itemize}[itemsep=2pt, topsep=0pt, parsep=0pt]
\item have the same semantic meaning as the "word to change".
\item keep the sentence's meaning unchanged.
\end{itemize}

Specifically, there are two scenarios for substitution:

\begin{itemize}[itemsep=2pt, topsep=0pt, parsep=0pt]
\item If the word to change is general, and we can clearly understand the sentence's meaning. In this case, the substitution should be more precise.
    (e.g. "Schools in north-west China are our primary aiding individuals and we often start from our school when the summer vacation begins." "aiding"→"helping" is a good substitution)
\item If the word to change is confusing, and we could only guess the sentence's meaning. In this case, the substitution should be more general.
    (e.g.  "Successful individuals are characterized by various merits including ..."  "various"→"plentiful" is a bad substitution)

\end{itemize}

After the substitution, the sentence must be fluent as the original sentence. Errors in preposition collocations, tenses, and mythologies should be avoided. (e.g. "in a nutshell", "nutshell" → "essence" is not right, should be "in a nutshell" → "in essence")

\subsection{Annotation Guidelines}

\begin{itemize}[itemsep=2pt, topsep=0pt, parsep=0pt]
\item Substitutions in a grid should be connected with ";" (NOT ',' !).
\item If the original sentence has grammar or typo problems, just discard the sentence.
\item In the annotation table, the content in the column "word to change" should be EXACTLY THE SAME as the word/phrase in the original sentence, and there should not exist punctuation (except ";" to connect multiple substitutions)
\item Substitute the smallest range of words, unless indivisible. (e.g. "I think you deserve it again" → "I think you deserve another chance" is a bad case, which should be "it again" → "another chance". "in a nutshell" → "in essence" is a good case, because "in a nutshell" is a phrase).
\item We don't need to paraphrase the sentence.
\item Please ensure that the "substitute" and "word to change" have the same tense, plural forms, and part of speech.
\end{itemize}

\section{Example of Evaluation Metrics} \label{evalexample}

For example, given a sentence: 
``I am writing to answer the previous questions you asked.''
The annotation result of the sentence is as follows:
$$
\begin{aligned}
& \mathtt{answer}\text{:\ }\mathtt{ \{respond\ to}\text{:\ }\mathtt{ 3,\ reply\ to}\text{:\ }\mathtt{ 1\}}, \\
& \mathtt{writing}\text{:\ }\mathtt{ \{connecting\ with}\text{:\ }\mathtt{ 3\}},\\
& \mathtt{to\ answer}\text{:\ }\mathtt{ \{in\ response\ to}\text{:\ }\mathtt{ 2\}},\\
& \mathtt{questions}\text{:\ }\mathtt{ \{queries}\text{:\ }\mathtt{ 2\}} \\
\end{aligned}
$$

In improvable target detection, $\text{S}_k$ is $\mathtt{\{answer,}$ $\mathtt{writing,\ to\ answer,\ questions\}}$. 
If the prediction $S'_k$ is $\mathtt{\{answer,\ previous\}}$, 
then $\mathbf{P^{det}} = 1 / 2$ and $\mathbf{R^{det}} = 1 / 4$. 

In substitution suggestion metrics, 
take the true predicted target $\mathtt{answer}$ as an example. 
If the predicted suggestion is in $\mathtt{\{respond\ to,}$ $\mathtt{reply\ to}$, 
then $\mathbf{Acc^{sug}} = 1$, 
otherwise $\mathbf{Acc^{sug}} = 0$. 

In end-to-end evaluation, 
if the predicted suggestions are $\mathtt{\{answer}\text{:\ }\mathtt{ respond,\ writing}\text{:\ }$ $\mathtt{ connect\ with,\ asked}\text{:\ }\mathtt{ gave\}}$, 
then $\mathbf{P^{e2e}} = 1 / 3$ and $\mathbf{R^{e2e}} = 1 / 4$.

\section{Prompt for ChatGPT} \label{chatgptprompt}

The prompt we use is as follows:

\texttt{In the following sentence, please give some suggestions to improve word usage. Please give the results with the json format of {``original word'': [``suggestion 1'', ``suggestion 2'']}, and the ``original word'' should be directly extracted from the sentence. [s]}

\noindent where \texttt{[s]} is the sentence. 
Amazingly, 
ChatGPT can generate substitution suggestions with the key-value format. 
We use regular expression to extract the substitution suggestions. 
If the result is empty, 
we will re-generate until getting substitution suggestions.

\section{Example of NDCG} \label{ndcgexample}

Take an example of $\mathbf{NDCG}_5$: 
For a detected improvable target, 
if $\text{T}_j$ with voting index is 
$\mathtt{\{respond\ to: 3,\ respond: 2,\ response: 1,}$ $\mathtt{reply\ to: 1\}}$ 
and $\text{T}'_j$ with order is 
$\mathtt{\{respond,}$ $\mathtt{respond\ to,\ tell,\ response,}$ $\mathtt{solution\}}$, 
then $\mathbf{DCG}(\text{T}_j')$ and $\mathbf{DCG}(\text{T}_j)$ are calculated as follows, and $\mathbf{NDCG}_5 = 4.4 / 5.1 = 86.3\%$.

\begin{table}[h]
\begin{tabular}{@{}llll@{}}
\toprule
Order & Sub.       & Gain & $\mathbf{DCG}_5(\text{T}'_k)$ \\ \midrule
1   & respond    & 2    & \ \ \  2 = 2 $\times$\ 1                  \\
2   & respond to & 3    & 3.9 = 2 + 3 $\times$\ 0.63         \\
3   & tell       & 0    & 3.9 = 3.9 + 0 $\times$\ 0.5        \\
4   & response   & 1    & 4.4 = 3.9 + 1 $\times$\ 0.43       \\
5   & solution   & 0    & 4.4 = 4.4 + 0 $\times$\ 0.39       \\ \midrule
Order & Sub.           & Gain & $\mathbf{DCG}_5(\text{T}_k)$ \\\midrule
1   & respond to     & 3    & \ \ \ 3 = 3 $\times$\ 1           \\
2   & respond        & 3    & 4.2 = 3 + 2 $\times$\ 0.63        \\
3   & response       & 1    & 4.7 = 4.2 + 1 $\times$\ 0.5       \\
4   & reply to       & 1    & 5.1 = 4.7 + 1 $\times$\ 0.43      \\
5   & NULL           & 0    & 5.1 = 5.1 + 0 $\times$\ 0.39      \\\bottomrule
\end{tabular}
\end{table}

\end{document}